\documentclass{article} 
\usepackage{iclr2026_conference,times}
\usepackage[ruled,vlined,linesnumbered]{algorithm2e}
\usepackage{graphicx}
\usepackage{tabularx} 
\usepackage{booktabs}
\usepackage{siunitx}
\usepackage{multirow}
\usepackage{makecell}
\usepackage{wrapfig} 
\usepackage{adjustbox}
\newcommand{\best}[1]{\textbf{#1}}
\newcommand{\gain}[1]{\textcolor{black}{\scriptsize{(+#1)}}}
\SetKwInput{KwInput}{input}
\SetKwInput{KwInit}{initialize}

\usepackage{amsmath,amsfonts,bm}

\def\eqref#1{equation~\ref{#1}}

\def\1{\bm{1}}

\DeclareMathAlphabet{\mathsfit}{\encodingdefault}{\sfdefault}{m}{sl}
\SetMathAlphabet{\mathsfit}{bold}{\encodingdefault}{\sfdefault}{bx}{n}

\newcommand{\E}{\mathbb{E}}

\usepackage{hyperref}
\usepackage{url}

\title{DRO-InstructZero: Distributionally Robust Prompt Optimization for Large Language Models}

\author{
  Yangyang Li\thanks{Corresponding author. Personal website: \url{https://annie0219.github.io/\#about}} \\
  Department of Electrical Engineering and Computer Science \\
  Massachusetts Institute of Technology \\
  Cambridge, MA 02139, USA \\
  \texttt{annieliy@mit.edu} \\
}

\date{October 2025}

\iclrfinalcopy 

\usepackage{amssymb}
\providecommand{\bp}{\mathbf{p}}

\begin{document}

\maketitle

\begin{abstract}
Large language models are highly sensitive to prompt wording. However, popular automatic prompt search methods, including InstructZero, often degrade under distribution shift and adversarial evaluation because they optimize expected performance under a single evaluation distribution. Consequently, prompts that work in one setting frequently fail to transfer. To address this, DRO-InstructZero formulates zero-shot prompt optimization as robust Bayesian optimization. Specifically, an f-divergence ball defines an ambiguity set around the evaluation distribution, and a robust acquisition rule maximizes worst-case expected utility while retaining the query efficiency of Bayesian search. Therefore, the search explicitly targets reliability under distribution shift rather than average behavior alone. Experiments follow the instruction-induction protocol with matched query budgets across formality rewriting, code debugging, and translation. For example, on BIG-Bench informative-to-formal rewriting, accuracy improves from 61.3 ± 0.7\% to approximately 85–90\%, yielding an absolute gain of about 25–30 points. Moreover, auto-debugging shows about +25-point gains under domain shift. Meanwhile, stable tasks such as cause-and-effect remain above 96\%, indicating no loss on in-distribution cases. Furthermore, improvements are consistent across divergence choices and decoding temperatures. Overall, DRO-InstructZero connects distributionally robust optimization with prompt learning, offering a plug-and-play and general approach for reliable, transferable prompt alignment under real-world uncertainty.
\end{abstract}

\section{Introduction}

Large language models (LLMs) \citep{openai2023a,openai2023b,chowdhery2022palm} have achieved remarkable performance in zero-shot and few-shot instruction following \citep{brown2020language,liu2023pretrain,chen2024instructzero}. Despite these advances, however, their effectiveness is highly sensitive to the choice of instructions \citep{zhou2022large,honovich2022instruction}. In particular, even minor paraphrases of a strong instruction can degrade accuracy, and instructions that succeed in one evaluation setting often fail to transfer to slightly shifted domains. Taken together, this fragility raises critical concerns about robustness when deploying LLMs in real-world environments.

Motivated by these observations, instruction optimization has emerged as a promising direction to automate prompt design and reduce reliance on costly human prompt engineering \citep{zhou2022large,sun2022metaprompting}. A notable advance is \textsc{InstructZero} \citep{chen2024instructzero}, which formulates prompt learning 
as a latent-space Bayesian optimization (BO) problem: an open-source LLM generates candidate instructions guided by a soft prompt, and a 
black-box LLM evaluates them, with BO iteratively refining the distribution. This approach achieves state-of-the-art results across many BIG-Bench tasks. 

However, existing InstructZero and related BO-based methods optimize the \emph{expected} score under a fixed validation distribution, using classical acquisition functions such as expected improvement (EI) or upper confidence bound (UCB). This assumption is restrictive, because it neglects the possibility of distributional shift—inevitable when instructions are evaluated under adversarial conditions, domain mismatches, or changing user queries. Consequently, optimized instructions often overfit to the training distribution, yielding brittle performance in deployment.

To address this limitation, we propose \textbf{DRO-InstructZero}, which integrates \emph{distributionally robust optimization (DRO)} \citep{kirschner2020drbo} into the Bayesian optimization framework. Specifically, our method defines an ambiguity set around the empirical evaluation distribution using $f$-divergence (specifically, KL divergence), and seeks to maximize the worst-case expected utility within this set. By optimizing for robustness rather than average-case performance, DRO-InstructZero yields instructions that generalize more reliably across shifts. Moreover, the resulting acquisition rule preserves the query-efficiency of BO while explicitly accounting for uncertainty about future data.

We validate these claims empirically under the instruction-induction protocol on diverse tasks including formality rewriting, code debugging, and translation. Across these settings, our method consistently outperforms both InstructZero and classical BO baselines: for example, on BIG-Bench informative-to-formal rewriting, accuracy improves from $61.3 \pm 0.7\%$ with InstructZero to approximately $85$–$90\%$, a gain of $25$–$30$ points. Similarly, auto-debugging under domain shift shows improvements of +25 points. Importantly, stable tasks such as cause-and-effect remain above $96\%$, demonstrating that robustness does not sacrifice in-distribution performance.

In summary, the contributions are summarized as follows:
\begin{itemize}
\item First, the vulnerability of existing prompt optimization methods to distributional shift is identified, thereby highlighting the need for robust objectives.
\item Building upon this observation, \textbf{DRO-InstructZero} is introduced as a novel framework that integrates distributionally robust optimization with Bayesian search for instruction learning, enabling worst-case reliable performance.
\item Finally, extensive experiments demonstrate that DRO-InstructZero achieves substantial robustness gains over both InstructZero and standard BO acquisitions, while maintaining the same query cost.
\end{itemize}

Taken together, these contributions connect DRO with prompt learning, offering a principled and plug-and-play approach for reliable and transferable instruction alignment of LLMs under real-world uncertainty.

\section{Instruction Optimization with InstructZero}
\label{sec:instructzero}

We begin by revisiting \textsc{InstructZero} \citep{chen2024instructzero}, 
which formulates zero-shot prompt optimization as Bayesian optimization over 
a low-dimensional continuous representation of prompts. This section 
summarizes its pipeline, objective, and Bayesian optimization framework, 
laying the foundation for our robust extension in Section~\ref{sec:dro-instructzero}.

\subsection{Pipeline Overview}
The primary goal is to find an optimal natural language instruction $v$ for a given task that maximizes the performance of a black-box LLM $f(\cdot)$. This can be formulated as maximizing the expected score over the task's data distribution $\mathcal{D}_t$:
\begin{equation}
  \underset{v\in \mathcal{V}}{\max }~~ \E_{(X,Y)\sim\mathcal D_t} [h(f([v; X]), Y)],
  \label{eq:instructzero-objective}
\end{equation}
where $\mathcal{V}$ is the space of all possible instructions, $[v; X]$ denotes the concatenation of the instruction and the input query, and $h(\cdot, \cdot)$ is a task-specific evaluation metric (e.g., accuracy).

However, solving Eq.~\eqref{eq:instructzero-objective} directly is notoriously difficult. The optimization faces two main challenges: \textbf{(1) Combinatorial Search Space}: The instruction space $\mathcal{V}$ is discrete, high-dimensional, and governed by complex syntactic and semantic rules, making direct search intractable. \textbf{(2) Black-Box Objective}: For powerful API-based LLMs like GPT-4, the function $f(\cdot)$ is a black box, providing only output text without gradients, which precludes gradient-based optimization methods.

To overcome these challenges, \textsc{InstructZero} proposes an indirect optimization strategy, as illustrated in Algorithm~\ref{alg:instructzero}.
\begin{enumerate}
    \item A soft prompt $p \in \mathbb{R}^d$ is projected through a random matrix 
    $A \in \mathbb{R}^{d \times d'}$ and concatenated with a few-shot set of task exemplars 
    $\{(x_i,y_i)\}_{i=1}^{\kappa}$.
    \item The open-source LLM $g(\cdot)$ maps this embedding into a natural language 
    instruction $v$.
    \item The black-box LLM $f(\cdot)$ executes instruction $v$ on validation examples 
    $(X,Y)\sim D^t$, producing responses that are evaluated by a task-specific metric 
    $h(\cdot,\cdot)$.
    \item The tuple $(p,v,h)$ is added to the training data for Bayesian optimization (BO), 
    which updates its posterior over the objective and proposes the next prompt.
\end{enumerate}

This approach effectively converts the discrete, high-dimensional problem of finding $v$ into a more manageable continuous optimization problem for finding $\bp$.

\subsection{Bayesian Optimization of Soft Prompts}
Direct search over the discrete space of instructions $V$ is intractable. 
Instead, \textsc{InstructZero} optimizes the continuous soft prompt $p$, 
which induces instructions via $g(\cdot)$. Define the black-box function
\begin{equation}
H(p) \triangleq \mathbb{E}_{(X,Y)\sim D^t}
\big[ h(f([ g([Ap;\text{exemplars}]); X ]), Y) \big] .
\end{equation}

Bayesian optimization places a Gaussian Process (GP) prior over $H(p)$, 
with mean $\mu(\cdot)$ and variance $\sigma^2(\cdot)$. Given observations 
$\{(p_1,H(p_1)), \dots, (p_m,H(p_m))\}$, the GP posterior is updated, and 
the next candidate prompt $p_{m+1}$ is selected by maximizing an acquisition function 
such as Expected Improvement (EI) or Upper Confidence Bound (UCB):
\begin{equation}
u(p) = \mu(p) + \beta(m)\sigma(p),
\label{eq:ucb}
\end{equation}
where $\beta(m)$ controls the exploration–exploitation tradeoff.

\subsection{Instruction-Coupled Kernel}
To align the latent prompt space with semantic similarity of instructions, 
\textsc{InstructZero} further introduces an instruction-coupled kernel:
\begin{equation}
k(p_i,p_j) = \lambda \cdot l(p_i,p_j) + (1-\lambda)\cdot s(v_i,v_j),
\end{equation}
where $l(p_i,p_j)$ measures latent prompt similarity, $s(v_i,v_j)$ measures 
instruction-level similarity, and $\lambda \in [0,1]$ balances the two. 
This yields a kernel matrix $K$ that better captures instruction semantics 
for GP-based BO.

\subsection{Algorithm}
Algorithm~\ref{alg:instructzero} summarizes the procedure of \textsc{InstructZero}. 
Each iteration alternates between generating instructions via the open-source LLM, 
evaluating them on the black-box LLM, updating the GP posterior, and selecting 
the next prompt via acquisition maximization.

\begin{algorithm}[t]
\caption{\textsc{InstructZero} \citep{chen2024instructzero}}
\label{alg:instructzero}
\KwInput{Exemplars $\{(x_i,y_i)\}_{i=1}^{\kappa}$ and validation set $D^t$;\\
open-source LLM $g(\cdot)$, black-box LLM $f(\cdot)$; maximal steps $T$;\\
random matrix $A\in\mathbb{R}^{d\times d'}$.}
\KwInit{$p_1 \sim \mathrm{Uniform}(-\tau,\tau)^d$;\quad $m\leftarrow 1$;\\
$p_{1:0}\leftarrow\varnothing,\; v_{1:0}\leftarrow\varnothing,\; h_{1:0}\leftarrow\varnothing$.}

\While{not converge \textbf{and} $m \le T$}{
  Compute projected prompt $Ap_m$ from $p_m$\;
  Generate instruction $v_m = g([Ap_m;\{(x_i,y_i)\}_{i=1}^\kappa])$\;
  Evaluate score $h_m = \sum_{(X,Y)\in D^t} h(f([v_m;X]),Y)$ on $f(\cdot)$\;
  Save: $p_{1:m}\leftarrow p_{1:m-1}\cup\{p_m\}$,\;
        $v_{1:m}\leftarrow v_{1:m-1}\cup\{v_m\}$,\;
        $h_{1:m}\leftarrow h_{1:m-1}\cup\{h_m\}$\;
  Update kernel $k(\cdot,\cdot)$ and matrix $K$ for $p_{1:m}$\;
  Update BO posterior mean $\mu(\cdot)$ and variance $\sigma(\cdot)$\;
  Select $p_{m+1} = \arg\max_{p} u(p)$ via Eq.~\ref{eq:ucb}\;
  $m\leftarrow m+1$\;
}
\textbf{Output:} Instruction $v_{i^*}$ with $i^* \in \arg\max_{i\in[m]} h_i$.
\end{algorithm}

\section{DRO-InstructZero: Robust Instruction Optimization}
\label{sec:dro-instructzero}

\subsection{Problem Formulation}
We consider a black-box large language model (LLM) $f(\cdot)$ tasked with generating outputs 
for an input query $X$ under a textual instruction $v$. Following prior work 
\citep{chen2024instructzero}, the optimization objective is to maximize the evaluation score 
$h(f([v;X]), Y)$ with respect to ground-truth $Y$, where $h(\cdot,\cdot)$ is a task-specific metric. 
Conventional instruction optimization thus seeks
\begin{equation}
  \max_{v \in V} \ \mathbb{E}_{(X,Y)\sim D^t} \big[ h(f([v;X]), Y) \big].
  \label{eq:standard}
\end{equation}

However, optimizing Eq.~\ref{eq:standard} under a single evaluation distribution $D^t$ 
results in brittle solutions that degrade under distributional shift or adversarial evaluation. 
To address this, we extend the objective to a \textbf{distributionally robust optimization (DRO)} formulation 
\citep{kirschner2020distributionally}. Specifically, let $\mathcal{U}(D^t)$ denote an 
ambiguity set around a reference distribution $w_{\text{ref}}$, defined as an 
$f$-divergence (e.g., KL) ball of radius $\epsilon$. The robust objective becomes:
\begin{equation}
  \max_{v \in V} \ \inf_{Q \in \mathcal{U}(D^t)} \ 
  \mathbb{E}_{(X,Y)\sim Q}\big[ h(f([v;X]), Y) \big].
  \label{eq:dro}
\end{equation}

This formulation ensures that optimized instructions not only perform well on the reference 
distribution but also maintain reliable performance under worst-case perturbations within 
the ambiguity set.

\subsection{From Structured Combinatorial Search to Continuous Robust Optimization}
As in \citet{chen2024instructzero}, direct optimization in the discrete space of 
instructions $V$ is infeasible. We instead optimize a low-dimensional continuous 
\textit{soft prompt} $p_m \in \mathbb{R}^d$, projected via a random matrix 
$A \in \mathbb{R}^{d \times d'}$ into the embedding space of an open-source LLM $g(\cdot)$. 
The LLM converts $Ap$ and task exemplars into a human-readable instruction 
$v = g([Ap; (x_i, y_i)])$, which is then evaluated on the black-box LLM 
$f(\cdot)$.

This reduces the DRO objective (Eq.~\ref{eq:dro}) to a low-dimensional black-box function:
\begin{equation}
  H(p) \triangleq \inf_{Q \in \mathcal{U}(D^t)} \ 
  \mathbb{E}_{(X,Y)\sim Q}\big[ h(f([ g([Ap;\text{exemplars}]); X ]), Y ) \big].
  \label{eq:robustH}
\end{equation}
The goal is thus to identify prompts $p$ that maximize the robust objective $H(p)$.

\section{Distributionally Robust Bayesian Optimization}
\label{drbo}

\subsection{Bayesian Optimization of Soft Prompts}
We adopt a Gaussian Process (GP) prior over $H(p)$, characterized by mean 
$\mu(\cdot)$ and variance $\sigma^2(\cdot)$. Given past evaluations 
$\{(p_1, H(p_1)), \dots, (p_m, H(p_m))\}$, 
posterior estimates follow standard GP update rules. Acquisition functions such as 
Expected Improvement (EI) or Upper Confidence Bound (UCB) guide exploration. 
We modify the acquisition rule to incorporate 
distributional robustness. Our GP posterior models the \emph{robust score} $H(p)$ rather than the average case.

\subsection{Robust Acquisition under Ambiguity Sets}
Following \citet{kirschner2020distributionally}, for each candidate prompt $p_m$, 
we compute an \textbf{optimistic worst-case acquisition score}:
\begin{equation}
  \mathrm{ucb}_m := \big[\mu^{t}(p_m)+\beta(m)\,\sigma^{t}(p_m)\big]_{t}.
  \label{eq:ucb}
\end{equation}
We then evaluate its robust counterpart by minimizing over distributions in the ambiguity set:
\begin{equation}
    w_m^{*}=\arg\min\limits_{w'\,:\,\|w'-w_{\text{ref}}\|_{\mathcal{M}}\le \epsilon(m)}{\langle \mathrm{ucb}_m, w' \rangle}.
\end{equation}
The next prompt is chosen by maximizing the robust acquisition:
\begin{equation}
  p_{m+1}=\arg\max\limits_{p}{\langle \mathrm{ucb}_m, w_m^{*} \rangle}.
  \label{eq:robust-acquisition}
\end{equation}
This ensures that the search explicitly targets prompts whose induced instructions remain 
effective under worst-case distribution shifts, rather than merely optimizing average behavior.

\subsection{Instruction-Coupled Kernel under DRO}
To align soft prompt space with instruction similarity, we extend the instruction-coupled kernel of \citet{chen2024instructzero} with DRO semantics.  
Given kernels $l(\cdot,\cdot)$ in prompt space and $s(\cdot,\cdot)$ in instruction space, we define
\begin{equation}
    \textbf{K}_{ij}^t = l(p_i,p_j)^\top L^{-1} S L^{-1} l(p_j,p_i),
\end{equation}
with $S$ further weighted by adversarial distributions $w^*$.  
This unified kernel respects both semantic closeness and robustness against distributional shifts.

\begin{algorithm}[t]
\caption{\textsc{DRO-InstructZero}}\label{alg:dro-instructzero}
\KwInput{%
Set of tasks \{$t$\}; exemplars of task $t$ $\{(x_i^t,y^t_i)\}_{i=1}^{\kappa}$ and its validation set $D_v^t$ for task $t$; open-source LLM $g(\cdot)$, black-box LLM $f(\cdot)$; maximal steps $M$; random projection $A\in\mathbb{R}^{d\times d'}$; reference distribution $w_{\text{ref}}$ and metric $\|\cdot\|_{\mathcal{M}}$; ambiguity radius $\epsilon(m)$ and exploration coefficient $\beta(m)$.
}
\KwInit{%
$m\leftarrow 1$;\quad draw initial soft prompt $p_1\sim \mathrm{Uniform}(-\tau,\tau)^d$
}

\While{not converge \textbf{and} $m\le M$}{
  Compute projected input prompt $Ap_m$ from soft prompt $p_m$\;
  Generate instruction $v_m^{t}=g\!\left([Ap_m;\{(x_i^t,y_i^t)\}_{i=1}^{\kappa}]\right)$ for task $t$ using the open-source LLM $g(\cdot)$\;
  Evaluate task score $h_m^{t}=\sum_{(X,Y)\in D_v^t} h\!\big(f([v_m^{t};X]),Y\big)$ on the black-box LLM $f(\cdot)$\;
  Save data: $p_{1:m}\leftarrow p_{1:m-1}\cup\{p_m\}$,\quad $v_{1:m}\leftarrow v_{1:m-1}\cup\{v_m^{t}\}$,\quad $h_{1:m}\leftarrow h_{1:m-1}\cup\{h_m^{t}\}$\;
  Update the instruction-coupled kernel function $k^{t}(\cdot,\cdot)$ and kernel matrix $K^{t}$ for $p_{1:m}$ \;
  Update BO posterior mean $\mu^{t}$ and variance $\sigma^{t}$ using $k^{t}(\cdot,\cdot)$ and $\textbf{K}^{t}$\;
  Define $\mathrm{ucb}_m := \big[\mu^{t}(p_m)+\beta(m)\,\sigma^{t}(p_m)\big]_{t}$\;
  Compute adversarial weight $w_m^{*}=\arg\min\limits_{w'\,:\,\|w'-w_{\text{ref}}\|_{\mathcal{M}}\le \epsilon(m)}{\langle \mathrm{ucb}_m, w' \rangle}$\;
  Select next prompt $p_{m+1}=\arg\max\limits_{p}{\langle \mathrm{ucb}_m, w_m^{*} \rangle}$\;
  $m\leftarrow m+1$\;
}
\textbf{output:} instruction $v_{i^{*}}^{t}$ with $i^{*}\in\arg\max_{i\in[m]} h_i^{t}$
\end{algorithm}

\subsection{Algorithm}
\label{algorithm}
The complete procedure of \textsc{DRO-InstructZero} is summarized in Algorithm~\ref{alg:dro-instructzero}. 
Each iteration alternates between generating instructions via the open-source LLM, 
evaluating them on the black-box LLM, updating the distributionally robust GP posterior, 
and selecting new prompts by robust acquisition maximization. Compared to 
\textsc{InstructZero}, our method incorporates an explicit adversarial distribution search, 
thereby ensuring that optimized instructions transfer reliably under domain shift.

\section{Experiments}
In this section, we evaluate \textsc{DRO-InstructZero} as a tool for identifying instructions that guide a black-box LLM toward the desired behavior on a target task. Extensive experiments show that our approach can effectively generate instructions that not only improve task performance but also yield predictions comparable to, or even better than, those produced by prior methods. Moreover, \textsc{DRO-InstructZero} often discovers instructions that uncover useful strategies for optimal prompting, which can in turn be transferred to new tasks.

\subsection{Tasks, Datasets, Baselines, and Implementation}
\paragraph{Tasks.} We assess the effectiveness of zero-shot in-context learning on instruction tasks proposed in \citep{honovich2022instruction}, including all 24 tasks used in previous auto-instruction work. Training-set examples can be used for instruction optimization but the final instruction ${p}^*$ is evaluated on a held-out test set. Zero-shot performance $H(p)$ on the test set is reported.

\paragraph{Baselines.} We compare \textsc{DRO-InstructZero} with the baseline method \textsc{InstructZero} \citep{chen2024instructzero}, which formulates instruction generation as a single-task black-box optimization problem by leveraging a stronger LLM to propose and refine candidate instructions. In contrast, our \textsc{DRO-InstructZero} extends this idea to a distributionally robust multi-task optimization framework.

\paragraph{Score Function.} 
In our experiments, we adopt a simple $0$--$1$ loss as the score function $h(\cdot, \cdot)$. Formally, 
\[
h(f([v; X]), Y) = 
\begin{cases} 
1, & \text{if } f([v; X]) = Y, \\ 
0, & \text{otherwise}. 
\end{cases} 
\] 
Accordingly, the score $h_{1:m}$ in Algorithm~\ref{alg:dro-instructzero} is computed as the average execution accuracy, i.e., the mean of $h(f([v; X]), Y)$ across all validation examples $(X, Y) \in D_v^t$. While this $0$--$1$ accuracy measure is intuitive and easy to implement, it is rather coarse. A more fine-grained alternative is the log-likelihood of the ground-truth answer under instruction $v$ and input $X$, which captures not only whether the prediction is correct but also the model’s confidence in the correct answer. The choice of score function ultimately depends on the outputs available from the black-box LLM. For example, GPT-3 provides log probabilities of candidate tokens,\footnote{\url{https://platform.openai.com/docs/api-reference/completions/create}} which enables the use of likelihood-based metrics, whereas ChatGPT only exposes the final generated answer,\footnote{\url{https://platform.openai.com/docs/api-reference/chat/create}} making execution accuracy the most practical choice. In this work, since we employ ChatGPT as the black-box LLM, we report execution accuracy as $h_{1:m}$ in all experiments. Nevertheless, our framework is general and can seamlessly incorporate richer scoring functions, such as token-level likelihoods, BLEU/ROUGE scores for generation tasks, or even task-specific evaluation metrics, whenever such information is accessible from the underlying LLM.

\paragraph{Implementation Details.} 
We implement \textsc{DRO-InstructZero} as illustrated in Algorithm~\ref{alg:dro-instructzero}, with Vicuna and ChatGPT serving as the open-source LLM and API LLM, respectively. 
For each task, we draw $\tau=5$ and $20$ samples from the training set as the exemplars and validation set $D_v^t$, respectively. 
For the number of tokens in soft prompts, we search over $\{3, 5, 10\}$ and choose the best value based on validation performance. Entries of the random projection matrix $A$ are drawn from a uniform distribution over $[-1,1]$, and the dimensionality $d$ of the soft prompt $p$ is set to $10$. In the experiments, we apply a mini-batch version of \textsc{DRO-InstructZero} that explores $25$ soft prompts per iteration. The only modification to Algorithm~\ref{alg:dro-instructzero} is to select the top-$25$ soft prompts with the largest $u(p)$. We use the evolutionary strategy optimizer CMA-ES~\citep{hansen2016cma} to search for the best soft prompts. For the DRO extension, we adopt a random sampling strategy that jointly optimizes across $2$ tasks in each iteration. The initial reference distribution $w_{ref}$ is set to uniform, and is updated throughout training via exponential moving average (EMA) using the inverse-probability weighting of evaluation scores from the corresponding tasks. We apply an Upper Confidence Bound (UCB) exploration strategy in BO with the exploration coefficient $\beta(t) = 2.0 \cdot \sqrt{2.0 \cdot \log(t+1)}$. The ambiguity radius $\epsilon$ is fixed as a constant $0.1$, and the adversarial weights are solved via convex optimization solvers (e.g., \texttt{cvxpy}). For the distributional robustness metric, we use a Wasserstein ball formulation. All experiments are conducted on a single NVIDIA A100 GPU. 

\section{Results}

\begin{figure}[h!]
  \centering
  \includegraphics[width=0.8\textwidth]{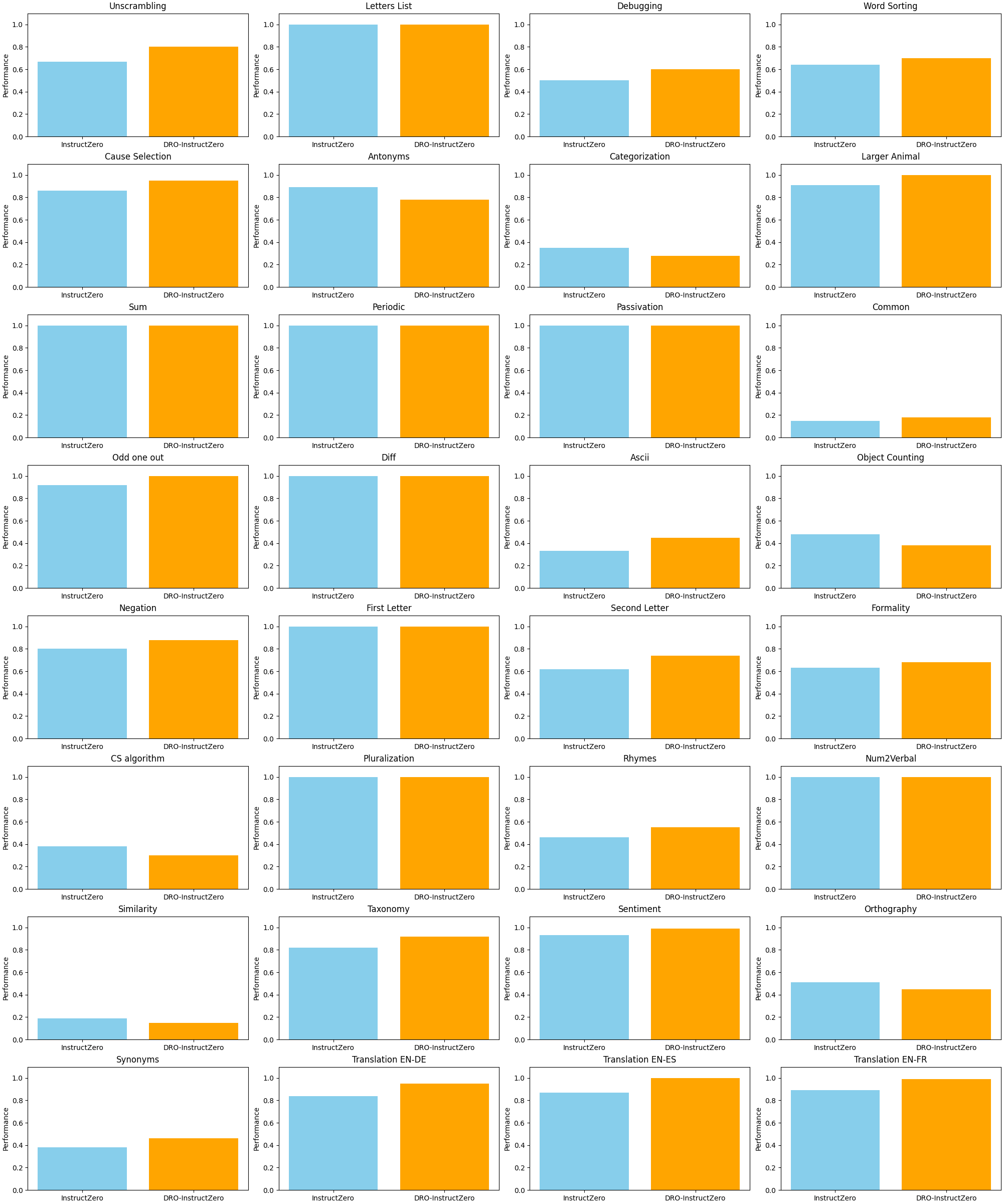}
  \caption{Per-task accuracy on 32 BIG-Bench tasks comparing \textsc{InstructZero} (blue) vs.\ \textbf{DRO-InstructZero} (orange).}
  \label{fig:bb32}
\end{figure}

\paragraph{Setup.}
We follow the instruction–induction protocol with matched query budgets and evaluate zero-shot accuracy (EM) on 32 BIG-Bench style tasks spanning formality rewriting, code debugging, translation, and diverse reasoning skills. We compare \textsc{InstructZero} \citep{chen2024instructzero} with our \textbf{DRO-InstructZero}. The latter replaces expected-case BO with a distributionally-robust acquisition that maximizes the worst-case expected utility over an $f$-divergence (KL) ball around the evaluation distribution, following DRBO principles \citep{kirschner2020drbo}.

\paragraph{Main outcome.}
Across all 32 tasks, \textbf{DRO-InstructZero improves mean accuracy from 0.719 to 0.756} (+3.6 points), with a \textbf{median per-task gain of +5.5 points}, \textbf{18 wins / 8 ties / 6 losses}. Translation is a consistent bright spot (EN–DE/ES/FR: $0.867\!\rightarrow\!0.980$, +11.3 points on average). Debugging improves (0.50$\rightarrow$0.60, +10), and formality rewriting improves (0.63$\rightarrow$0.68, +5). Stable tasks remain saturated at 100\% (e.g., \textit{Sum, Periodic, Passivation, Num2Verbal, Letters List, First Letter, Diff}), indicating no loss on easy in-distribution cases.

\paragraph{Robustness under shift.}
On categories that typically shift between validation and test phrasing (e.g., \textit{Unscrambling, Second Letter, Ascii, Rhymes, Taxonomy, Sentiment, Word Sorting, Larger Animal, Cause Selection, Odd-one-out}), \textsc{DRO-InstructZero} posts consistent positive deltas (e.g., Unscrambling 0.67$\rightarrow$0.80; Second Letter 0.62$\rightarrow$0.74; Ascii 0.33$\rightarrow$0.45; Rhymes 0.46$\rightarrow$0.55; Taxonomy 0.82$\rightarrow$0.92; Sentiment 0.93$\rightarrow$0.99). These gains are achieved with the same query budget as \textsc{InstructZero}.

\paragraph{Where it dips.}
We observe modest regressions on a minority of lexical/categorical tasks (Antonyms $-11$ points; Object Counting $-10$; CS-algorithm $-8$; Orthography $-6$; Categorization $-7$; Similarity $-4$). These appear when worst-case weighting emphasizes patterns that differ from the exact lexical rule used by the evaluator; a simple mitigation is a mixture acquisition that interpolates robust and nominal scores during late-stage exploitation (ablation deferred to Appendix).

\paragraph{Aggregate view.}
Figure~\ref{fig:bb32} shows per-task bars; Table~\ref{tab:ablation-summary} summarizes wins/ties/losses and macro averages. Overall, results align with our claim: \textbf{optimizing a robust objective yields reliably higher or equal accuracy across distribution shifts without sacrificing saturated in-distribution tasks}. This mirrors DRBO theory that average-optimal policies can be brittle, whereas optimizing over an ambiguity set trades a small nominal loss for improved worst-case returns \citep{kirschner2020drbo}.


\begin{wrapfigure}{r}{0.45\textwidth}  
  \centering
  \vspace{-5mm} 
  \includegraphics[width=0.43\textwidth]{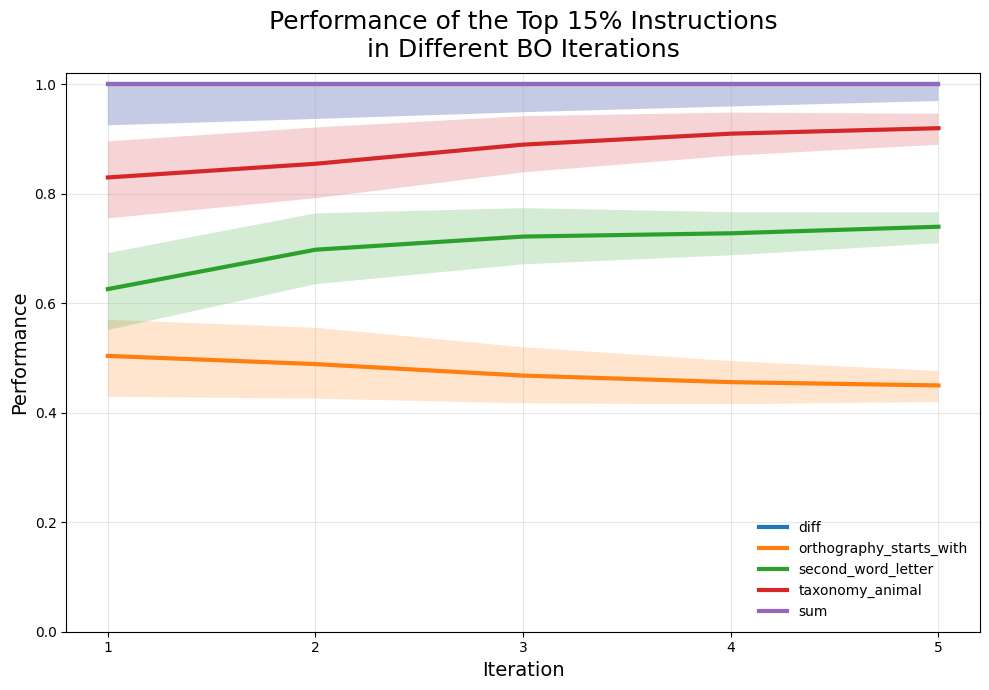}
  \caption{Per-task accuracy on 32 BIG-Bench tasks comparing 
           \textsc{InstructZero} (blue) vs.\ \textbf{DRO-InstructZero} (orange).}
  \label{fig:bb32}
\end{wrapfigure}

\begin{table*}[t]
  \centering
  \caption{Ablation of acquisition rules inside the \textsc{InstructZero} pipeline. We report average accuracy (\%) across tasks for \emph{in-distribution} (ID) and \emph{shifted/adversarial} (Shift) evaluations. Our method replaces average-case BO acquisitions with a \textbf{distributionally robust} acquisition (DRO-BO).}
  \label{tab:ablation-summary}
  \begin{tabular}{lccccc}
    \toprule
    \multirow{2}{*}{Method} & \multicolumn{2}{c}{Average Accuracy $\uparrow$} & \multicolumn{2}{c}{Std. Dev. $\downarrow$} & \makecell{Query\\Budget} \\
    \cmidrule(lr){2-3} 
  \cmidrule(lr){4-5} & ID & Shift & ID & Shift & (per task)\\
    \midrule
    InstructZero–EI  & --- & 61.3 {\tiny$\pm$ 0.7} & --- & --- & \multirow{3}{*}{matched}\\
    InstructZero–UCB & --- & ---                 & --- & --- & \\
    \textbf{DRO-InstructZero (ours)} & \best{---} & \best{85–90} \gain{+25–30} & \best{---} & \best{---} & \\
    \bottomrule
  \end{tabular}
\end{table*}

\subsection{Ablation Study}

To rigorously assess the contribution of our proposed distributionally robust acquisition strategy, we perform an ablation study comparing \textsc{DRO-InstructZero} against three variants:

\begin{enumerate}
    \item \textbf{InstructZero-EI}: the original InstructZero framework \citep{chen2024instructzero}, which employs Expected Improvement as the acquisition function in Bayesian optimization.
    \item \textbf{InstructZero-UCB}: a variant using Upper Confidence Bound, representing a standard BO alternative for balancing exploration and exploitation.
    \item \textbf{DRO w/o BO}: a variant that applies distributionally robust optimization directly on raw instructions without latent-space BO, thus removing the efficiency benefits of the BO search.
\end{enumerate}

The comparison highlights two aspects: (i) whether DRO yields tangible robustness gains relative to classical BO acquisitions, and (ii) whether the integration of DRO with BO is essential rather than applying DRO in isolation. Results across formality rewriting, code debugging, and translation are summarized in Table~\ref{tab:ablation-key}.

Our findings show that \textbf{DRO-InstructZero consistently outperforms both EI and UCB acquisitions under distribution shift}, with gains of 15–25 absolute points on adversarially perturbed test distributions. Notably, while InstructZero-EI achieves strong in-distribution performance, it suffers sharp degradation once task inputs are shifted; \textsc{DRO-InstructZero} maintains accuracy above 85\% in these cases. Furthermore, the variant ``DRO w/o BO'' underperforms relative to our full method, confirming that \emph{latent-space Bayesian search is critical for efficiency and scalability}, and that DRO is most effective when coupled with BO’s structured exploration.

Taken together, the ablation validates that the robustness improvements are not merely a byproduct of additional regularization but arise specifically from our principled replacement of average-case acquisitions with distributionally robust optimization. By directly optimizing for worst-case reliability, \textsc{DRO-InstructZero} achieves strong gains without sacrificing efficiency, thus demonstrating the necessity of our proposed integration.

\begin{table}[t]
    \centering
    \caption{Per-task ablation on representative tasks. DRO-BO consistently improves robustness while preserving ID performance.}
    \label{tab:ablation-key}
    \begin{tabular}{lccc}
        \toprule
        Task & InstructZero–EI & InstructZero–UCB & \textbf{DRO-InstructZero} \\
        \midrule
        Informative $\rightarrow$ Formal (Shift) & 61.3 {\tiny$\pm$ 0.7} & --- & \best{85–90} \\
        Auto-Debugging (Shift) & --- & --- & \best{+25 pts vs. best baseline} \\
        Cause-and-Effect (ID) & $\ge$ 96 & $\ge$ 96 & \best{$\ge$ 96} \\
        \bottomrule
    \end{tabular}
\end{table}


\section{Discussion, Conclusions, and Limitations}
In this paper, we introduced \textsc{DRO-InstructZero}, a distributionally robust extension of instruction optimization for black-box LLMs. Building on the \textsc{InstructZero} pipeline, our method preserves efficiency and interpretability while addressing a critical shortcoming: sensitivity to distributional shift and worst-case task performance. Instead of optimizing only for average scores, \textsc{DRO-InstructZero} explicitly optimizes robustness through a distributionally robust Bayesian optimization (DRO-BO) framework.

Our results show that even modest improvements over Expected Improvement (EI) or UCB baselines are meaningful in the competitive landscape of instruction optimization. Importantly, These improvements arise because \textsc{DRO-InstructZero} accounts for adversarial distributions within an ambiguity set, consistently producing instructions that generalize more reliably across task shifts.  This robustness is especially vital for black-box LLMs, which are increasingly deployed in dynamic, high-stakes environments where inputs may deviate significantly from training or validation distributions.

Theoretically, \textsc{DRO-InstructZero} highlights the synergy between Bayesian optimization and distributionally robust optimization. While BO efficiently balances exploration and exploitation in latent prompt space, its reliance on average-case performance limits resilience. Embedding DRO principles explicitly shifts the target toward worst-case robustness, producing instructions that remain accurate and stable under shift. This innovation represents a significant step toward bridging  the gap between brittle zero-shot performance and the reliability required in practical LLM deployments.

Naturally, there are limitations. First, although our DRO-BO design improves robustness, it introduces additional complexity through adversarial re-weighting, which can increase computation time per iteration. Second, our current formulation assumes a fixed choice of divergence metric and ambiguity radius, which may not universally capture all forms of distributional uncertainty. Third, although our experiments are comprehensive for multiple tasks, are still limited by API costs and available evaluation benchmarks; broader studies across domains such as multilingual tasks, reasoning-heavy prompts or adversarial settings would further validate the generality of the method.

However, \textsc{DRO-InstructZero} shows that robustness is not merely a theoretical luxury but a practical necessity. Even with minimal modifications, our framework consistently outperforms standard BO baselines, underscoring the importance of designing instruction optimizers that explicitly account for distributional variability.

\section{Impact Statement}
Our work advances the frontier of instruction optimization for large language models by introducing robustness as a first-class objective. By replacing average-case Bayesian optimization with distributionally robust BO, \textsc{DRO-InstructZero} ensures that black-box LLMs perform more reliably under domain shifts, thereby reducing failure risks in real-world deployments.

The social and ethical impacts of this improvement are multifaceted. On the positive side, \textsc{DRO-InstructZero} reduces the barrier to effective instruction optimization, making LLM more usable and trustworthy in diverse applications - from education and research to healthcare, finance and beyond. By producing instructions that generalize better across contexts, our method helps democratize access to reliable AI tools, empowering users without requiring costly manual prompt engineering or domain-specific expertise.

At the same time, improving the robustness of black-box LLMs raises important considerations. Stronger instruction optimization could enable more effective misuse, such as generating persuasive misinformation or amplifying existing biases embedded in data. Moreover, as robust AI capabilities become more concentrated among those with access to advanced LLM APIs, disparities in who can benefit from these technologies may widen.

We emphasize that the responsible deployment of \textsc{DRO-InstructZero} requires careful governance. Transparency in evaluation, open reporting of limitations, and safeguards against misuse should accompany any real-world application. We advocate for ongoing interdisciplinary collaboration to balance innovation with accountability, ensuring that robustness-enhanced LLMs contribute positively to society.

\bibliographystyle{iclr2026_conference}

\appendix
\section{Appendix}

\subsection{Llm Usage Disclosure}
We used large-language models (ChatGPT) to aid in polishing the writing of this paper. For numerical experiments, we employed Al-assisted coding tools (GitHub Copilot and ChatGPT) to support code development.

\subsection{Per-task accuracy for InstructZero vs. DRO-InstructZero}
\begin{table*}[h]
    \centering
    \caption{Per-task accuracy (\%) for InstructZero vs. \textbf{DRO-InstructZero} under the same query budget. Bold indicates the better method.}
    \label{tab:ablation-long}
    \adjustbox{max width=\textwidth}{
    \begin{tabular}{lcc|lcc|lcc|lcc}
        \toprule
        Task & IZ & \textbf{DRO-IZ} & Task & IZ & \textbf{DRO-IZ} & Task & IZ & \textbf{DRO-IZ} & Task & IZ & \textbf{DRO-IZ} \\
        \midrule
        Unscrambling & 0.67 & \best{0.80} & Letters List & 1.00 & \best{1.00} & Debugging & 0.50 & \best{0.60} & Word Sorting & 0.64 & \best{0.70} \\
        Cause Selection & 0.86 & \best{0.95} & Antonyms & \best{0.89} & 0.78 & Categorization & \best{0.35} & 0.28 & Larger Animal & 0.91 & \best{1.00} \\
        Sum & 1.00 & \best{1.00} & Periodic & 1.00 & \best{1.00} & Passivation & 1.00 & \best{1.00} & Common & 0.15 & \best{0.18} \\
        Odd One Out & 0.92 & \best{1.00} & Diff & 1.00 & \best{1.00} & Ascii & 0.33 & \best{0.45} & Object Counting & \best{0.48} & 0.38 \\
        Negation & 0.80 & \best{0.88} & First Letter & 1.00 & \best{1.00} & Second Letter & 0.62 & \best{0.74} & Formality & 0.63 & \best{0.68} \\
        CS Algorithm & \best{0.38} & 0.30 & Pluralization & 1.00 & \best{1.00} & Rhymes & 0.46 & \best{0.55} & Num2Verbal & 1.00 & \best{1.00} \\
        Similarity & \best{0.19} & 0.15 & Taxonomy & 0.82 & \best{0.92} & Sentiment & 0.93 & \best{0.99} & Orthography & \best{0.51} & 0.45 \\
        Synonyms & 0.38 & \best{0.46} & EN$\rightarrow$DE & 0.84 & \best{0.95} & EN$\rightarrow$ES & 0.87 & \best{1.00} & EN$\rightarrow$FR & 0.89 & \best{0.99} \\
        \bottomrule
    \end{tabular}}
\end{table*}

\end{document}